\documentclass[twoside,twocolumn]{article}

\usepackage{tabulary,graphicx,times,caption,fancyhdr,amsfonts,amssymb,amsbsy,latexsym,amsmath}
\usepackage[utf8]{inputenc}
\usepackage{url,multirow,morefloats,floatflt,cancel,tfrupee,textcomp,colortbl,xcolor,pifont}
\usepackage[nointegrals]{wasysym}
\makeatletter

\usepackage{ifxetex}
\ifxetex\else
  \usepackage{dblfloatfix}
\fi

\@ifundefined{subparagraph}{
\def\subparagraph{\@startsection{paragraph}{5}{2\parindent}{0ex plus 0.1ex minus 0.1ex}%
{0ex}{\normalfont\small\itshape}}%
}{}

\def\URL#1#2{\@ifundefined{href}{#2}{\href{#1}{#2}}}

\def\UrlOrds{\do\*\do\-\do\~\do\'\do\"\do\-}%
\g@addto@macro{\UrlBreaks}{\UrlOrds}

\makeatother

\usepackage[paperheight=11in,paperwidth=8.3in,margin=2.5cm,headsep=.7cm,top=2.5cm]{geometry}
\usepackage[T1]{fontenc}

\usepackage{array}
\usepackage{color}
\usepackage{xcolor}
\usepackage{amsmath}
\usepackage{amsmath,amssymb,amsfonts}
\usepackage[noend]{algpseudocode}

\usepackage{graphicx}
\usepackage{caption}
\usepackage{subcaption}
\usepackage{textcomp}
\usepackage{longtable}
\usepackage{enumitem}
\usepackage{rotating}
\usepackage{hyperref}
\usepackage{amsxtra}
\usepackage{amssymb}
\usepackage{latexsym}
\usepackage{amsmath,amsfonts,amssymb,amsthm}
\usepackage{multirow}
\usepackage{amsmath,amssymb,amsfonts}
\usepackage{graphicx}
\usepackage{textcomp}
\usepackage{algorithm}
\usepackage[noend]{algpseudocode}

\renewenvironment{abstract}
	{\trivlist\item[]\leftskip0pt\par\vskip4pt\noindent
  	\textbf{\abstractname}\mbox{\null}\\}
	{\par\noindent\endtrivlist}

\def\keywords#1{\par\medskip\par\noindent\textbf{Keywords}: #1\par}

 \date{} \emergencystretch 8pt

\makeatletter
\def\author#1{\gdef\@author{\hskip-\tabcolsep%
	\parbox{\textwidth}{\raggedright\bfseries#1\\[1pc]}}}
\def\address[#1]#2{\g@addto@macro\@author{\\\hskip-\tabcolsep\parbox{\textwidth}{\raggedright%
	\normalsize\normalfont\textsuperscript{#1}#2}}}
\let\addresslink\textsuperscript
\def\correspondence#1{\g@addto@macro\@author{\\\hskip-\tabcolsep\parbox{\textwidth}{\raggedright%
	\vspace*{10pt}\normalsize\normalfont~\\#1~\\[12pt]}}}
\def\email#1{\g@addto@macro\@author{\\\hskip-\tabcolsep\parbox{\textwidth}{\raggedright%
	\normalsize\normalfont Emails: #1}}}

\def\title#1{\gdef\@title{\vspace*{-30pt}
	\raggedright\textbf{Computational Intelligence and Neuroscience}~\\%
  \raggedright\bfseries\ifx\@articleType\@empty\vspace*{20pt}\else%
  \vspace*{20pt}\@articleType\vspace*{20pt}\\\fi#1}}
\let\@journaltitle\@empty \def\journaltitle#1{\gdef\@journaltitle{{\normalfont\itshape#1}}}
\let\@articleType\@empty \def\articletype#1{\gdef\@articleType{{\normalfont\itshape#1}}}

\let\@runningHead\@empty \def\RunningHead#1{\gdef\@runningHead{{\normalfont #1}}}

\usepackage{fancyhdr}
\fancypagestyle{headings}{\fancyhf{}
  \fancyhead[R]{\itshape\@runningHead}
  \fancyfoot[C]{\thepage}}
\fancypagestyle{plain}{%
	\fancyhf{}\fancyhead[R]{\LaTeX\ template for Hindawi journal articles}
  \fancyfoot[C]{\thepage}}
\makeatother

\usepackage[%
	numbers,sort&compress%
  ]{natbib}

\usepackage{float,xcolor}

\journaltitle{<journal title>}
\articletype{<Research Article>} 

\begin{document}

\title{\textcolor{black}{Genetic CFL: Hyper-Parameter Optimization in Clustered Federated Learning}}

\author{%
		Shaashwat Agrawal\addresslink{1},
  	Sagnik Sarkar\addresslink{1},
  	Mamoun Alazab\addresslink{2},
  	Praveen Kumar Reddy Maddikunta\addresslink{3},
  	Thippa Reddy Gadekallu*\addresslink{3} and
  	Quoc-Viet Pham\addresslink{4}
     }
		
\address[1]{School of Computer Science and Engineering, VIT, Vellore, 632014, India}
\address[2]{College of Engineering, IT and Environment, Charles Darwin University, Casuarina, NT 0909, Australia}
\address[3]{School of Information Technology and Engineering, VIT, Vellore, 632014, India}
\address[4]{Korean Southeast Center for the 4th Industrial Revolution Leader Education, Pusan National University, Busan 46241, Korea}

\correspondence{Thippa Reddy Gadekallu: thippareddy.g@vit.ac.in}

\email{shaashwat.agrawal2018@vitstudent.ac.in(Shaashwat Agrawal), sagnik.sarkar2018@vitstudent.ac.in(Sagnik Sarkar), alazab.m@ieee.org(Mamoun Alazab), praveenkumarreddy@vit.ac.in(Praveen Kumar Reddy Maddikunta), thippareddy.g@vit.ac.in(Thippa Reddy Gadekallu), vietpq@pusan.ac.kr(Quoc-Viet Pham)}%

\RunningHead{Running head}

\maketitle 
\footnote{https://github.com/sagnik106/Clustered-FL-GA}
\begin{abstract}
 Federated learning (FL) is a distributed model for deep learning that integrates client-server architecture, edge computing, and real-time intelligence. FL has the capability of revolutionizing machine learning (ML) but lacks in the practicality of implementation due to technological limitations, communication overhead, non-IID (independent and identically distributed) data, and privacy concerns. Training a ML model over heterogeneous non-IID data highly degrades the convergence rate and performance. The existing traditional and clustered FL algorithms exhibit two main limitations, including inefficient client training and static hyper-parameter utilization. To overcome these limitations, we propose a novel hybrid algorithm, namely genetic clustered FL (Genetic CFL), that clusters edge devices based on the training hyper-parameters and genetically modifies the parameters cluster-wise. Then, we introduce an algorithm that drastically increases the individual cluster accuracy by integrating the density-based clustering and genetic hyper-parameter optimization. The results are bench-marked using MNIST handwritten digit dataset and the CIFAR-10 dataset. The proposed genetic CFL shows significant improvements and works well with realistic cases of non-IID and ambiguous data. \textcolor{black}{An accuracy of 99.79\% is observed in the MNIST dataset and 76.88\% in CIFAR-10 dataset with only 10 training rounds.}   

\end{abstract}

\section{Introduction}

Federated Learning (FL) \cite{zhang2021survey,pham2021fusion} has risen as a groundbreaking sub-domain of Machine Learning (ML) that enables Internet of Things (IoT) devices to contribute their real-time data and processing to train ML models. FL represents a distributed architecture of a central server and heterogeneous clients, aiming to reduce the empirical loss of model prediction over non-independent and identically distributed (non-IID) data. In contrast to traditional ML algorithms that require large amounts of homogeneous data in a central location, FL utilizes on-device intelligence over distributed data \cite{alazab2021federated,wang2021secure}. The limited feasibility of ML in industrial and IoT applications is overturned by the introduction of FL. Some potential applications of FL include google keyboard \cite{yang2018applied}, image-based geolocation \cite{sprague2018asynchronous}, healthcare informatics \cite{xu2020federated}, and wireless communications \cite{pham2021uav}. 

Each round of a FL paradigm constitutes of client-server communication, local training and model aggregation \cite{nilsson2018performance}. The communication overhead is usually due to model broadcast from the server to all clients and vice-versa. In every communication round, there is a feasibility risk in terms of limited network bandwidth, packet transmission loss, and privacy breach. In the growing applications of Industrial Internet of Things (IIoT), where the communication is Machine to Machine (M2M) these parameters may be static, making efficiency in data transfer important. Modified communication algorithms \cite{fang2021privacy} use compression and encryption to reduce the model size and protect privacy. Communication load is also determined by the number of edge devices. Sparsification of communication \cite{ozfatura2021time} implemented over clients is modeled to increase convergence rate and reduce network traffic on the server. Many models also utilize hierarchical clustering \cite{inproceedings} to generalize similar client models and reduce the aggregation complexity. 

Apart from communication, training ML models in a heterogeneous setup presents a huge challenge \cite{zhao2018federated}. Once the server model is broadcast, the clients train on it considering some hyper-parameters such as client ratio (i.e., from a strength of 100, number of clients chosen), learning rate, batch size, epochs per round, etc. With each edge device, computational power and properties of data (ambiguity, size, complexity) vary drastically and diversely trained client models are hard to aggregate. In a realistic scenario of thousands of edge devices, the updated global model may not converge at all. Existing aggregating algorithms like FedAvg and FedMA \cite{wang2020federated} focus more on integration of weights of the local models. Convergence rate and learning saturation are common concerns when it comes to training and aggregation. Several novel approaches work around model aggregation either by using feature fusion of global and local models \cite{yao2019towards} or by a grouping of similar client models \cite{ghosh2020efficient} to increase generalization. Some literatures also utilize multiple global models to better converge data \cite{kopparapu2020fedcd}.

\begin{figure*}[h!]
	\centering
	\includegraphics[width=1.\linewidth]{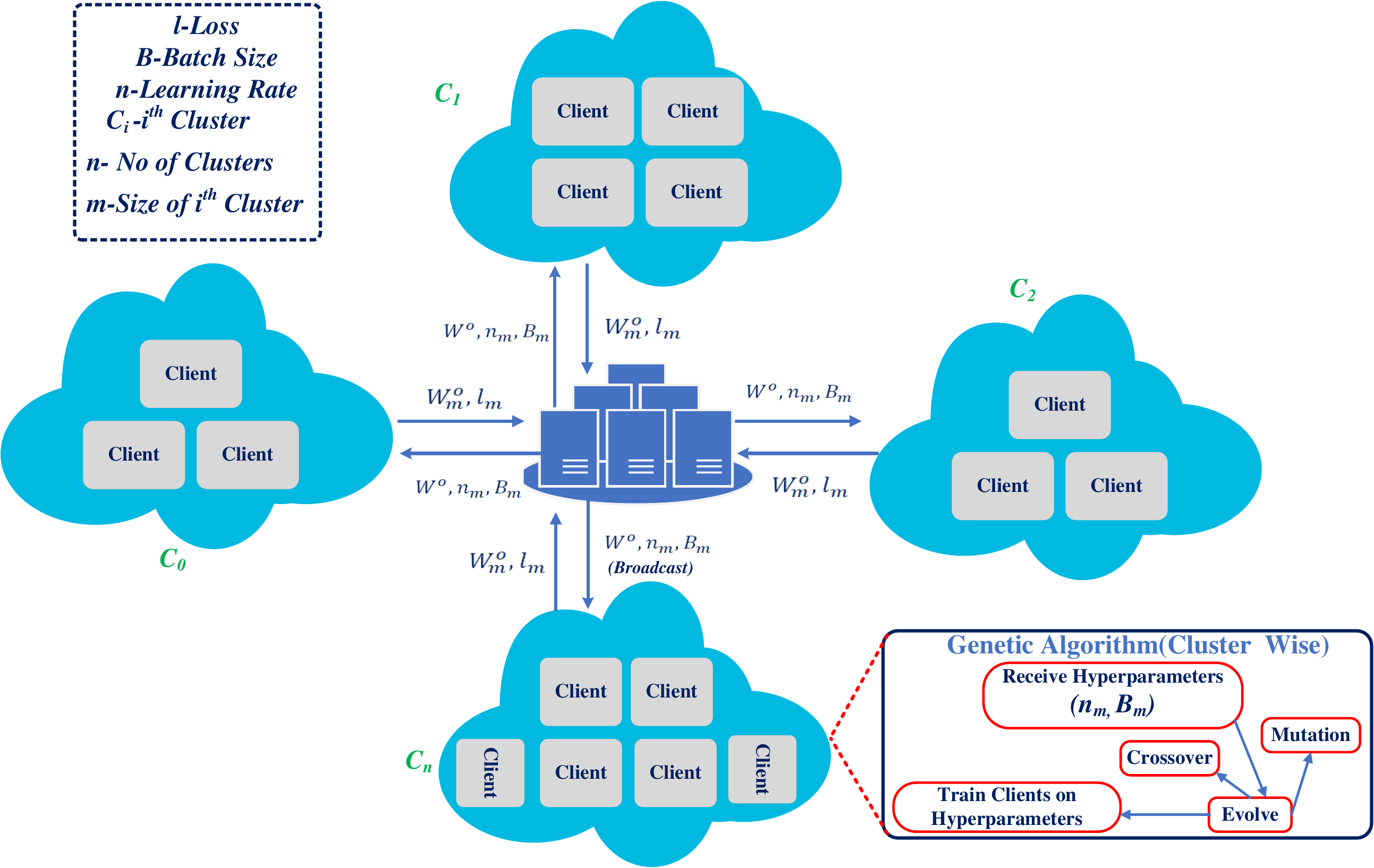}
	\caption{\textcolor{black}{Genetic CFL complete architecture}}
	\label{Fig:algo}
\end{figure*}

Research on making FL models adaptive to non-IID data has focused primarily on model aggregation. Local training of the model itself is an undermined step, given its role in the final accuracy. In this paper, we propose three novel contributions to lessen the empirical risk in FL as shown in Fig.~\ref{Fig:algo}:  
\begin{itemize}
  \item Clustering of clients solely based on model hyper-parameters to increase the learning efficiency per unit training of model.
  \item Implementation of density-based clustering, i.e., DBSCAN, on the hyper-parameters for proper analysis of devices properties.
  \item Introduction of genetic evolution of hyper-parameters per cluster for finer tuning of individual device models and better aggregation.
\end{itemize}
In particular, we introduce a new algorithm, namely Genetic CFL, that clusters hyper-parameters of a model to drastically increase the adaptability of FL in realistic environments. Hyper-parameters such as batch size and learning rate are core features of any MFL model. In truth, every model is tuned manually depending on its behavior to the data. Therefore, in a realistic heterogeneous setup, the proper selection of these parameters could result in significantly better results. DBSCAN algorithm is used since it is not deterministic, static in terms of cluster size and uses neighbourhood of model hyper-parameters for clustering. We also introduce genetic optimization of those parameters for each cluster. \textcolor{black}{Genetic Algorithm is algorithm since it is highly application flexible and scalable to to higher dimensions.} As defined, each cluster of clients has its own unique set of properties(i.e. hyper-parameters) that are suitable for the training of the respective models. In each round, we determine the best parameters for each cluster and evolve them to better suit the cluster. 

The rest of this paper is organized as follows. \textcolor{black}{The Related Work Section \ref{relatedwork} discusses the recent work done in the fields of FL, Clustering, and Evolutionary optimization algorithms. The proposed algorithm is defined in the Genetic CFL Architecture Section~\ref{completealgo}, followed by the results in Experiments and Results~\ref{results}. Finally, the paper is concluded in Conclusion~\ref{sec:conclusion}.}

\section{Related Work} \label{relatedwork}

\textcolor{black}{In this section, we survey the current literature on the topics of FL, Density-Based Clustering, Evolutionary Algorithms respectively and try to understand their limitations. } 

\subsection{Federated Learning} \label{lit_fl}

Recently, FL as a distributed and edge ML architecture is being studied extensively \cite{zhang2021survey, bonawitz2019towards}. This decentralized nature of FL directly contradicts traditional ML algorithms which are genuinely difficult to train in a heterogeneous environment consisting of non-IID data. Novel approaches have tried to overcome this difficulty through various model aggregation algorithms namely FedMA \cite{wang2020federated}, feature fusion of global and local models \cite{yao2019towards}, agnostic FL and grouping of similar client models \cite{ghosh2020efficient} for better personalization and accuracy. Clustering takes advantage of data similarity in various clients and models \cite{xie2020multi}, efficient communication, and lastly improves global generalization \cite{chai2020tifl}. In general, much work is yet to be done in terms of efficient model training on non-IID data.

\subsection{Density-based Clustering} \label{lit_dbc}

Clustering in FL is primarily used for efficient communication and better generalization. In a realistic scenario with thousands of nodes, aggregating everything into a single model may damp the convergence greatly. Several partitioning, hierarchical and density-based clustering algorithms have been applied to work on some of the problems existing in FL. Partitioning clustering algorithms like k-means clustering \cite{likas2003global} demand a predetermined number of clusters, but in actuality that is not feasible. Some example of non-definitive clusters include Agglomorative Hierarchical clustering \cite{briggs2020federated} and generative adversarial network based clustering. In this paper, we propose to use DBSCAN (Density-Based Spatial Clustering of Applications with Noise) \cite{birant2007st}, a density-based clustering algorithm that only groups points if they satisfy a density condition.                                                              

\subsection{Evolutionary Algorithms} \label{lit_ea}

Hyper-parameters of a model determines their ability to learn from a certain set of data. Optimization of ML models and their hyper-parameters using evolutionary algorithms \cite{kim2019evolutionary} like Whale Optimization \cite{aljarah2018optimizing} and Genetic Algorithms \cite{xiao2020efficient} are explored by many researchers. \textcolor{black}{In addition, these algorithms have been extensively used over DL frameworks that have become a trend for optimization tasks \cite{beruvides2014online}. The same has yet not been adopted for FL extensively. Also, algorithms such as Reinforcement Learning (RL) with focus on Q-Learning are not suitable for highly complex scenarios \cite{beruvides2015self}.} The need for hyper-parameter tuning increases even more in FL due to the ambiguity in data and the above-mentioned optimization algorithms assist in tuning those parameters beyond manual capacity. Since optimization of each client model parameters is not feasible, we propose to do so for each cluster. 

Through the survey, we observe that FL is greatly limited by efficiency of individual client training that includes apt choice of hyper-parameters, increasing adaptive nature of the models and optimization of such process.

\section{Genetic CFL Architecture} \label{completealgo}

In this section, we give a detailed mathematical model of our algorithm, Genetic CFL. The complete pipeline is divided into two parts, the initial broadcast round represented by Algorithm~\ref{alg:algo1} to determine the clusters and the federated training using genetic optimization represented by Algorithm~\ref{alg:algo2}. The variational behavior of the algorithm with different hyper-parameters, including client ratio($n$), number of rounds, $\epsilon$, minimum samples, learning rate($\eta$), and batch size is explained in this section. Table \ref{symbols} elucidates all the symbols utilizedd in the algorithm

\begin{table}[htbp]
\caption{\textcolor{black}{Symbol Representations}}
\begin{center}
{\color{black}\begin{tabular}{|c|c|}
\hline
\textbf{Symbol} & \textbf{Meaning}\\
\hline
n & number of clients\\
\hline
$\eta$ & learning rate\\
\hline
B & batch size\\
\hline
C$_i$ & i$^{th}$ client\\
\hline
w$^0$ & model weights \\
\hline
w$^0_n$ & model weight of n$^{th}$ client\\
\hline
\end{tabular}}
\label{symbols}
\end{center}
\end{table}

\begin{algorithm} 
\caption{Initial Broadcast Round and Clustering\newline
n = Number of Clients\newline
$\eta$: learning rate\newline
$\eta_{m}$: learning rate list}
\label{alg:algo1}
\begin{algorithmic}[1]

\Procedure{Server}{}    
    \State $w^0$: Server Model Initialization
    \State Initialize $\eta_{m}$ with learning rates $[1e-1, 1e-6]$
    \State Broadcast($w^0$, random($\eta_{m},3$))
\EndProcedure
\Procedure{Client}{}
    \State $i \gets 0$
    \While{$i \not= n$}  
        \State $j \gets 0$
        \While{$j \not= 3$}
            \State Train $w_{j}^{0}$ on $\eta_j$
            \State $losses \gets$ loss($w^0_j$)
            \State \textcolor{black}{j = j + 1}
        \EndWhile  \label{client loop}
        \State \textcolor{black}{i = i + 1}
        \State \Return{$w_{min}^0, \eta_{min}, losses_{min}$} 
    \EndWhile  \label{learning rate loop}
\EndProcedure
\Procedure{Server}{}
    \State $w^0 \gets \frac{\sum(w^0_n)}{n}$    
    \State Initialize DBSCAN Clustering Algorithm
    \State $\epsilon \gets 1e-6$ 
    \State $points_{min} \gets 2$
    \State $model \gets$ DBSCAN$(\epsilon, points_{min})$
    \State clusters = model$.$fit\_predict($\eta_n$)
\EndProcedure
\end{algorithmic}
\end{algorithm}

The purpose of Algorithm~\ref{alg:algo1} is to discreetly determine the data characteristics of an edge device without intruding on their privacy. A server model($w^0$) is initialized and broadcast to $n$ clients, C $\subseteq$ \{$C_0, C_1,..., C_{tot}$\}. With each distributed model, three different $\eta$ are broadcast. The sample size is chosen to introduce variance in training, while more number of samples can also be used for experiments. These learning rates are chosen from an array($\eta_m$) ranging from $[1e-1,1e-5]$. Each edge device receives $w^0$ that is cloned for all values of $\eta$ and trained individually for a single epoch. Data properties unique to an edge device like size, complexity, ambiguity, and variance drastically affect the training, and thus hyper-parameters of a model: $\eta$, batch size are chosen accordingly. Naturally, from the three trained models in an edge device, the one with the least loss, denoted as $w^0_{\min}$, is chosen. Each edge device then returns $w^0_{\min}$, $\eta_{\min}$, and $losses_{\min}$. The significance of these values is their data representative capacity of the respective edge devices.

At server, the models $\{w^0_0, w^0_1, \dots, w^0_n\}$, the learning rates $\{\eta_0, \eta_1, \dots, \eta_n\}$ and their respective losses are attained. The model aggregation technique is used to obtain the server model by combining edge device models. The weights of the models ($w^0_n$) are summed iteratively as follows:
\begin{equation}
    \sum\limits_{i=0}^{n}(w^0_i) = w^0_0 + w^0_1 +  w^0_2 + \dots + w^0_n. \label{eq:summ}
\end{equation}
After summation, the output of the equation divided by the number of clients to obtain model aggregation as
\begin{equation}
    w^0 \gets \frac{\sum(w^0_n)}{n}. \label{eq:avg}
\end{equation}

After server model aggregation, the DBSCAN  clustering algorithm is applied. In a realistic scenario, the number of edge devices and their variance cannot always be determined. In deterministic partitioning clustering methods like K-Means clustering, the number of clusters have to be predetermined and is not dynamic. DBSCAN, on the other hand, uses density-based reasoning for the grouping of similar objects. It takes two mandatory inputs, $\epsilon$ and min samples. Any point $x$ forms a cluster if a minimum number of samples lie in its $\epsilon-$ neighbourhood. This value can be calculated by
\begin{align}
    & A_{\epsilon} \equiv \{(x \in HP_{space}), dist(x,y) < \epsilon \}, \label{eq:epsilon}\\
    & HP_{space}(\eta) \in [1e-3, 1e-7], \label{eq:lr_bound}\\
    & HP_{space}(batch-size) \in [16, 128]. \label{eq:bs_bound}
\end{align}
Here, $HP_{space}$ represents the domain in which the point $x$ must be presented. In our case, it is the range of hyper-parameters, specifically learning rate $\eta$. Each $\epsilon$-neighbourhood must contain a certain number of points (MinPts) to be called a cluster as follows:
\begin{equation}
     A_{\epsilon} > MinPts. \label{eq:cluster_cond}
\end{equation}

\begin{equation}
     MinPts \in \mathbb{N}\label{eq:pts}
\end{equation}
In the object space of only learning rate, $\left|\eta_{2}-\eta_{1}\right|$ gives the Euclidean distance used for $\epsilon$ neighbourhood. When the number of dimensions is increased with the addition of batch size(B) the eucladian distance formula for 2 coordinate system is used and logarithmic values of hyper parameters are taken to scale the exponential values to liner ones. The calculation can be observed as

\begin{equation}
     \epsilon = \sqrt{\left(\log_{10}\left(\frac{\eta_2}{\eta_1}\right)\right)^2 + \left(\log_2\left(\frac{B_2}{B_1}\right)\right)^2} \label{eq:eu_dist}
\end{equation}
After each edge device is allotted a cluster-ID, we implement phase-2, shown by Algorithm \ref{alg:algo2}. This section of the algorithm works under the main control loop which runs for $i$ rounds. Every $i$-th iteration, 
\begin{enumerate}
    \item Hyper-Parameters are optimized per cluster using Genetic Algorithm involving evolution followed by crossover and finally mutation. 
    \item Server model with optimized hyper-parameters are broadcast to each client cluster wise.
    \item Each client is trained based on said parameters.
    \item Client models are aggregated to form the latest server model.
\end{enumerate}

\begin{algorithm}[h!]
\caption{Genetic Optimization based FL on clustered data.\newline
rounds: Number of loops for training the federated model}
\label{alg:algo2}
\begin{algorithmic}[1]
\Function{Mutate}{$\eta$}
  \State factor $\gets$ random($[-1,0,1]$)
  \State $\eta \gets$ $\eta + \frac{\eta\times{factor}}{10}$

    \State \Return {$\eta$}
\EndFunction

\Function{Crossover}{$\eta_n$}
  \State Initialize temporary array $\eta_{temp}$ to store $\eta$
  \State $\eta_{temp}[0,1] \gets \eta_n[0,1]$
      \For{$k \gets 2$ to $size(\eta_n)$}    
        \State $parent_A, parent_B \gets$ random($0, size(\eta_n)$)
        \State $\eta_{temp}[k] \gets \Call{Mutate}{\frac{\eta_n[parent_A] + \eta_n[parent_B]}{2}}$
    \EndFor

    \State \Return {$\eta_{temp}$}
\EndFunction
\Function{Evolve}{$losses_n, \eta_n$}
  \State $losses_n$, order $\gets$ sort($losses_n$)
  \State $\eta_n \gets$ sort($\eta_n$, order)

    \State \Return {\Call{Crossover}{$\eta_n$}}
\EndFunction
\Procedure{Train}{}   
    \State len $\gets$ size(clusters)
    \State ind $\gets$ $0$ to len
    \State Initialize $\eta_{global}$ with shape (len, size(clusters$[$ind$])$
    \State clusters$_{unique}$ = unique(cluster)
      \For{$i \gets 0$ to rounds}    
          \For{$k \gets 0$ to $size($clusters$)$}    
                \State ind = [clusters.index(cluster[i])]
                \State $\eta_{global}[k] \gets \Call{Evolve}{$losses[ind],$\eta_n[ind}$
        \EndFor
        \State Empty arrays losses, $\eta_n$
        \For{$k \gets 0$ to n}    
                \State $w^0_k[k],$ losses[k], $\eta_n[k]$ = train($w^0$, $\eta_{global}[$clusters[k][clusters[k].nextIndex])
                
        \EndFor
        \State $w^0 \gets \frac{\sum(w^0_m)}{n}$    
        
    \EndFor
\EndProcedure

\end{algorithmic}
\end{algorithm}

Every cluster has a different set of characteristic hyper-parameters suitable to the edge devices belonging to them. These clustered parameters are evolved genetically followed by training for every $i$-th round. Using Genetic Optimization for tuning converges the set of hyper-parameters to an optimal set each round. $\eta_{global}[k]$ is initialized that stores learning rates for each cluster and its contents are modified every round. It is of shape $<C, size(C_{i})>$, where $C$ is the number of clusters, $C_i$ represents the $i_{th}$ cluster and $size(C_{i})$ represents the number of edge devices in each $i_{th}$ cluster. The hyper-parameters of a cluster having shape $m_0$ are sorted through their losses. 
\begin{align}
    & losses_n, order \gets sort(losses_n) \label{eq: sort_ord} \\
    & \eta_n \gets sort(\eta_n, order) \label{eq: sort}
\end{align}
Once sorted, we obtain new individuals through crossover and mutation respectively. The best individuals (hyper-parameters in a cluster) retain their genes and are promoted to the next generation (round) while the others are formed by mating of individuals from the last generation as
\begin{equation}
     \eta_{new} = \{\eta_0, \eta_1, \dots, \eta_{size(C[m])}\} \label{eq: new}
\end{equation}
The new learning rates $\eta_{new}$ are chosen either directly or by mating. The number of $\eta$ taking from old generation can vary. From \ref{eq: new} we derive the modified parameters:
\begin{equation}
     \eta_{new} = \{\eta_{old}[0], \eta_{old}[1], \dots, \left(\frac{\eta_{old}[P_A] + \eta_{old}[P_B]}{2}\right)\left(1+\frac{f}{10}\right)\}, \label{eq: new_gen}
\end{equation}
where $P_A, P_B \in [0, 9]$ and $f \in [-1, 1]$.

After genetic evolution, the server model is again broadcast to all devices with their respective cluster hyper-parameters. Each edge device trains for 1 epoch and the complete process of genetic optimization, training, and model aggregation is repeated for $i-1$ rounds. 

\section{Experiments and Results} \label{results}

This section deals with the experiments that have been conducted to validate and test the proposed Genetic CFL architecture. \textcolor{black}{The following subsection \ref{sec:dbscan} deals with the clustering of the client edge devices and the clustering behavior under various parameters. Sections after DBSCAN \ref{sec:perform_cifar}, \ref{sec:perform_mnist} is concerned with the performance of the Genetic CFL architecture on MNIST and CIFAR-10 datasets respectively and their comparison with the generic FL architecture. The overall performance analysis for the Genetic CFL architecture is discussed in the final subsection performance analysis\ref{sec:perform_analys}.}

\subsection{DBSCAN Clustering of the client models}\label{sec:dbscan}
The DBSCAN algorithm, as discussed in the previous section, focuses on the Euclidean distance between the observations to calculate the density and cluster the observations based on this density. The models in each edge device is assigned a particular learning rate and batch size for training. These two hyper-parameters serve as the primary two dimensions for each observation for the process of clustering. The DBSCAN algorithm takes two main parameters for clustering a set of observations: $\epsilon$ and \textit{Min Samples}. We note that $\epsilon$ is the maximum Euclidean distance for an observation from the closest point in the cluster in question. The \textit{Min Samples} parameter is the least number of observations possible in the clustering algorithm. Thus the tuning and selection of these parameters become essential to obtain proper and efficient results.

\begin{table}[t]
\caption{DBSCAN clustering parameters and outcomes}
\begin{center}
\begin{tabular}{|c|c|c|}
\hline
$\epsilon$ & \textbf{\textit{Min Samples}} & \textbf{Number of clusters}\\
\hline
\multirow{2}{*}{0.200} 
& 1 & 7\\
\cline{2-3}
& 2 & 7\\
\hline
\multirow{2}{*}{0.175} 
& 1 & 7\\
\cline{2-3}
& 2 & 7\\
\hline
\multirow{2}{*}{0.150}
& 1 & 8\\
\cline{2-3}
& 2 & 7\\
\hline
\multirow{2}{*}{0.100}
& 1 & 15\\
\cline{2-3}
& 2 & 18\\
\hline
\end{tabular}
\label{dbscan}
\end{center}
\end{table}

\begin{figure}[h!]
\centering
\includegraphics[width=1.\linewidth]{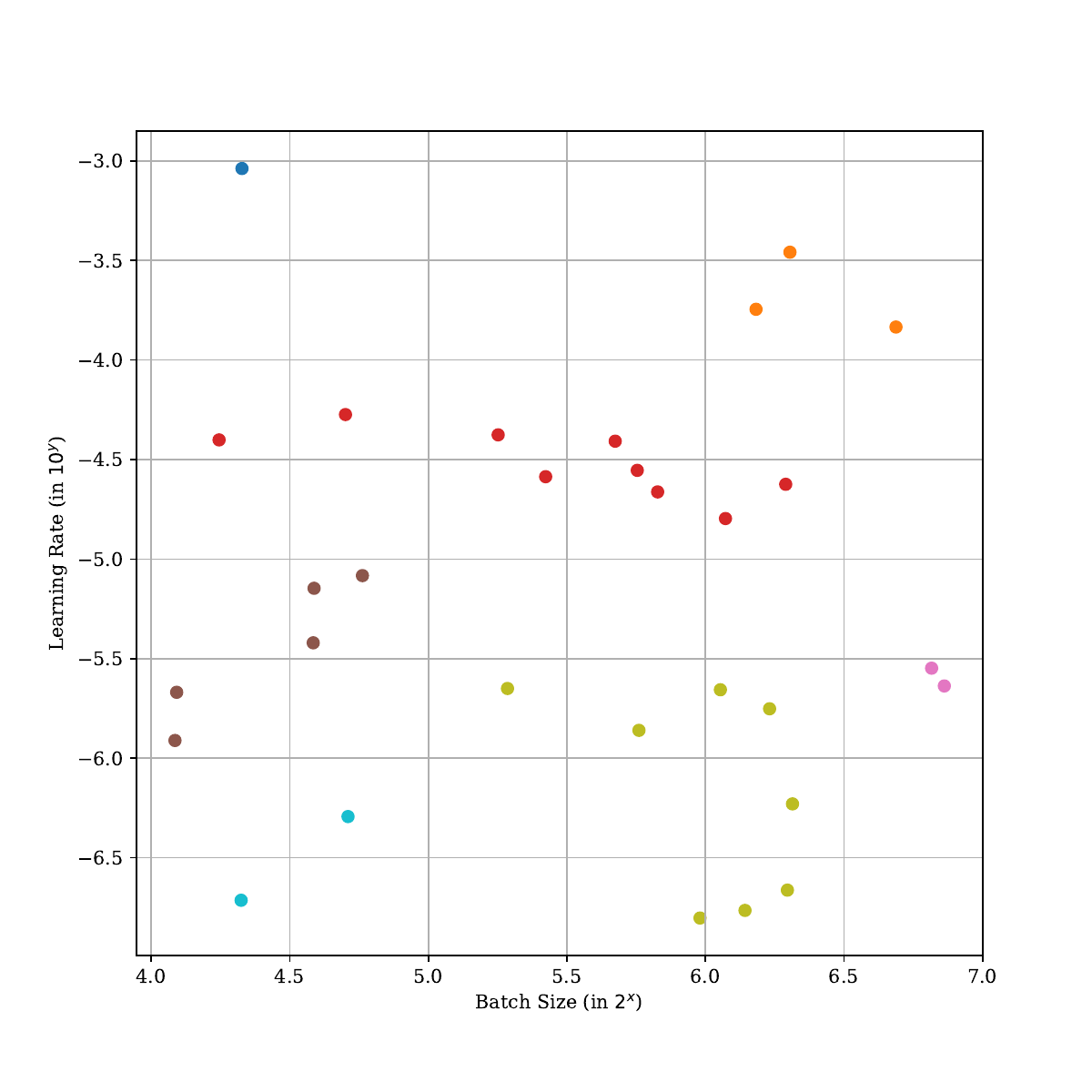}
\caption{Clustering of client devices based on the learning rate and batch size by using DBSCAN algorithm.}
\label{fig:cluster}
\end{figure}

\begin{figure}[h!]
\centering
\includegraphics[width=1.\linewidth]{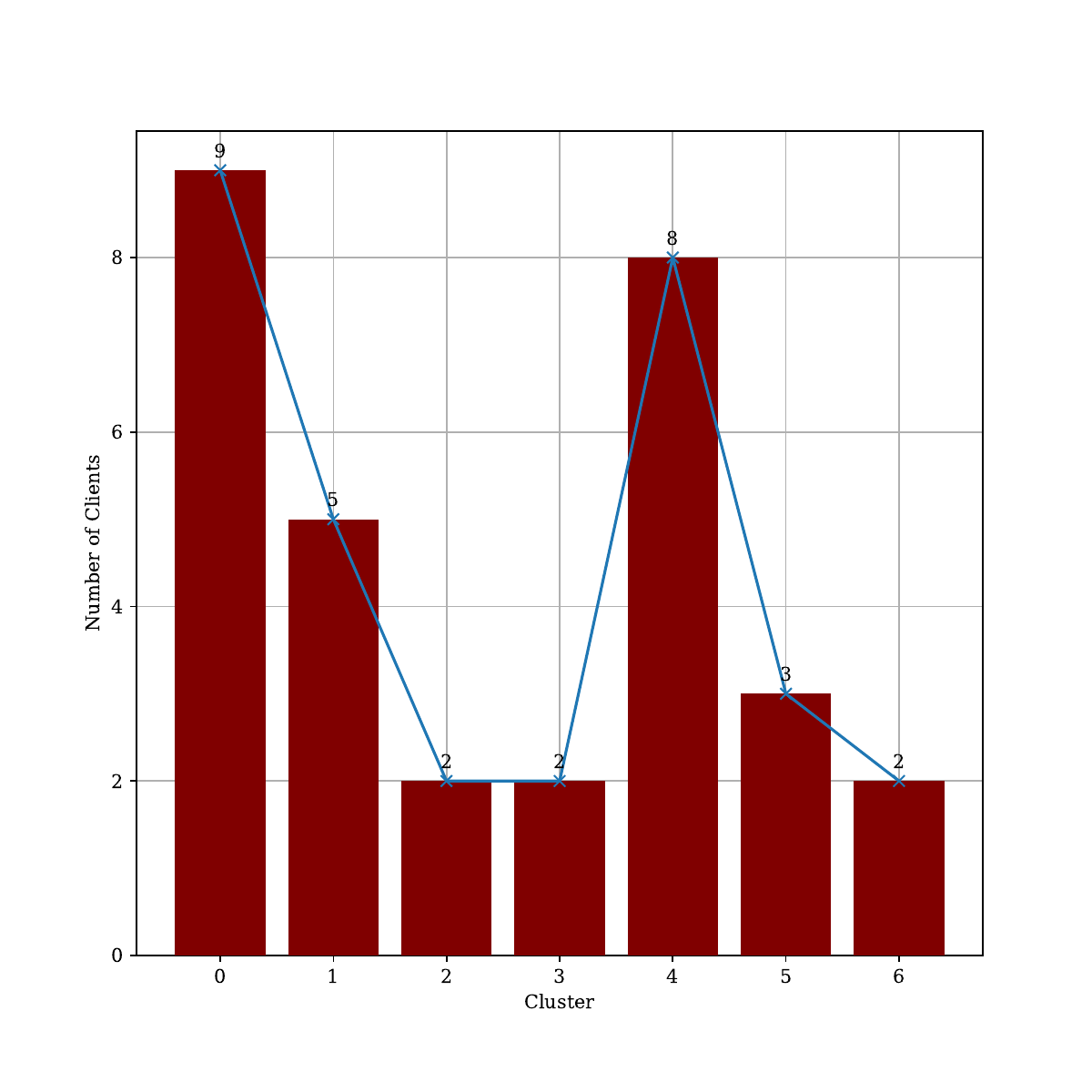}
\caption{Number of observations in each cluster.}
\label{fig:clust_no}
\end{figure}

Table~\ref{dbscan} summarizes the conditions tested for the quality and effectiveness of clustering with the said parameters. For each value of $\epsilon$ two values of \textit{Min Samples} are tested to validate the clustering effectiveness and detecting outliers in the data. For the $\epsilon$ values $0.2$ and $0.175$ the number of clusters for both 1 and 2 \textit{Min Samples} stay constant at 7. This constant value for the generated number of clusters for both the \textit{Min Samples} indicate that there are no outliers in the data and each observation in the cluster holds a strong relationship with each other. Since the number of clusters for both the epsilon values are the same, it is evident that the clusters are locally isolated. For the $\epsilon$ values $0.150$ and $0.100$, the number of clusters change drastically indicating weak clustering among the observations. The change in number of clusters for different \textit{Min Samples} is proof that there are outliers in the data which can cause issues while performing the genetic optimization due to the lack of population. The parameters can therefore be safely assigned either of the four combinations to obtain 7 distinct clusters as shown in Fig.~\ref{fig:cluster}. The number of observations in each cluster is plotted in Fig.~\ref{fig:clust_no}.

\subsection{Performance of the Genetic CFL architecture on MNIST dataset}\label{sec:perform_mnist}

\begin{figure*}[htbp]
\includegraphics[width=1\linewidth]{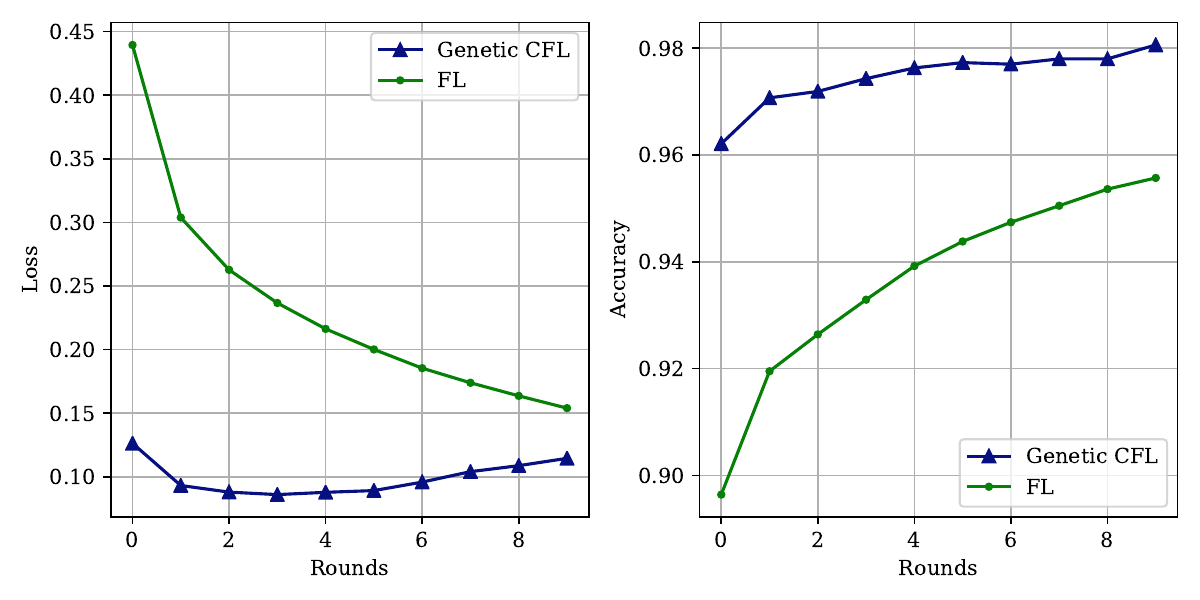}
\caption{Round-wise performance of the FL Architecture compared with Genetic CFL architecture on MNIST Handwritten Digit Dataset}
\label{fig:mnist_per}
\end{figure*}

\textcolor{black}{In this subsection we discuss the performance and analyse the training curves of the models. The server model is initially trained on a subset of the MNIST handwritten digits dataset \cite{deng2012mnist}}. This model is then distributed among the clients based on the client ratio. The total number of clients chosen for this experiment is $100$ and the client ratios tested for are $0.1$, $0.15$ and $0.3$. In essence, we evaluate the performance of the models on $10$, $15$ and $30$ clients respectively. Each client device is provided with a random subset of the dataset with a random number of observations. This is to make sure that the data is non-IID and the characteristics of the real time scenario is emulated. For the initial round, the hyper-parameters (learning rate and batch size) of the client devices are randomized within the intervals \eqref{eq:lr_bound} and \eqref{eq:bs_bound} respectively. The client devices are trained for two epochs and the hyper-parameters are subjected to genetic evolution as discussed in Section~\ref{completealgo}. These rounds are tabulated in Table~\ref{mnist_cfl} and the best performance is plotted against each round in Fig.~\ref{fig:mnist_per}.

\begin{table*}[t]
\caption{Performance of FL against Genetic CFL on MNIST dataset for various hyper-parameters.}
\begin{center}
\begin{tabular}{|c|c|c|c|c|c|c|}
\hline
\multirow{2}{*}{\textbf{Client Ratio}}&\multirow{2}{*}{\textbf{Rounds}}&\multicolumn{2}{|c|}{\textbf{FL}}&\multicolumn{2}{|c|}{\textbf{Genetic CFL}} \\
\cline{3-6} 
& & \textbf{\textit{Accuracy}}& \textbf{\textit{Loss}} & \textbf{\textit{Accuracy}}& \textbf{\textit{Loss}}\\
\hline
\multirow{3}{*}{0.1} & 3 & 0.9133 & 0.3136 & 0.9679 & 0.1203\\
\cline{2-6}
& 6 & 0.9265 & 0.2493 & 0.9730 & 0.1343\\
\cline{2-6}
& 10 & 0.9367 & 0.2115 & 0.9777 & 0.1923\\
\hline
\multirow{3}{*}{0.15} & 3 & 0.9176 & 0.2878 & 0.9665 & 0.1049\\
\cline{2-6}
& 6 & 0.9740 & 0.0876 & 0.9740 & 0.0876\\
\cline{2-6}
& 10 & 0.9443 & 0.1828 & 0.9763 & 0.0910\\
\hline
\multirow{3}{*}{0.3} & 3 & 0.9178 & 0.2989 & 0.9698 & 0.0964\\
\cline{2-6}
& 6 & 0.9326 & 0.2359 & 0.9780 & 0.0804\\
\cline{2-6}
& 10 & 0.9450 & 0.1946 & 0.9799 & 0.0849\\
\hline
\end{tabular}
\label{mnist_cfl}
\end{center}
\end{table*}

Since the model training hyper-parameters are no longer predetermined, the performance of the models and their respective training are optimized locally in the cluster, thus providing a more personalized training for each cluster. The performance of the server model obtains a smooth learning curve and converges faster than the normal training of the model using FL. Table~\ref{mnist_cfl} represents this performance of the models for both the architectures. The superiority of performance of Genetic CFL over Generic FL is evident for each round. The accuracy of the Genetic CFL architecture is consistently higher and the loss is consistently lower as compared to the generic FL architecture. The increase in accuracy and the decrease in loss signify that the models are indeed training and useful information is aggregated at the server.

\subsection{Performance of the Genetic CFL architecture on CIFAR10 dataset}\label{sec:perform_cifar}

\begin{figure*}[h!]
\includegraphics[width=1\linewidth]{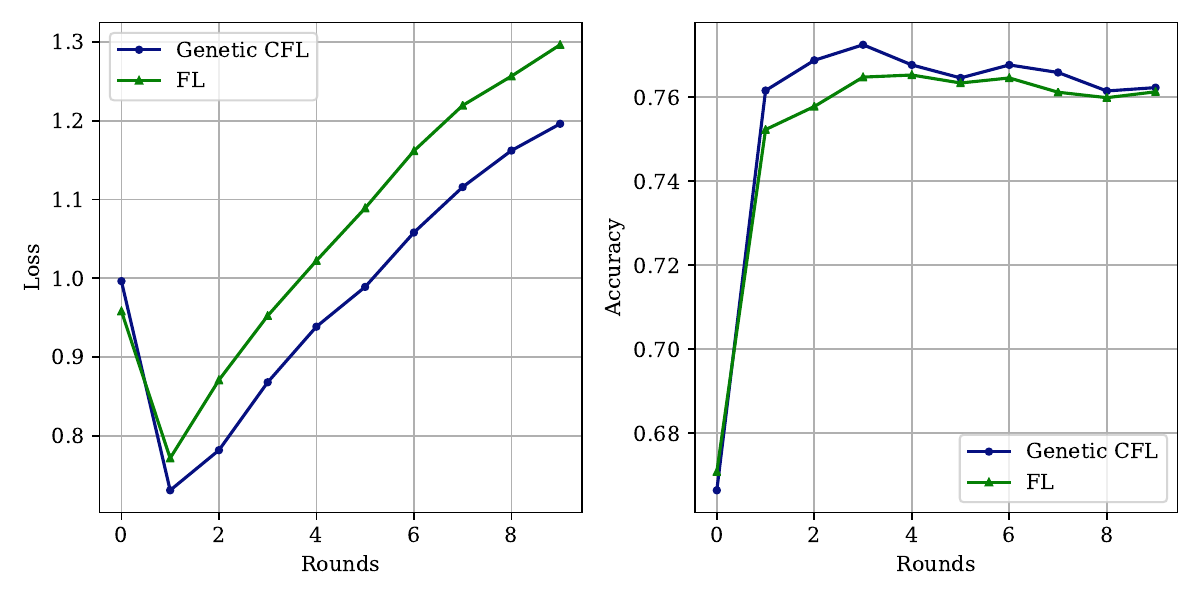}
\caption{Round-wise performance of the FL Architecture compared with Genetic CFL architecture on CIFAR-10 Dataset.}
\label{fig:cifar_per}
\end{figure*}

\textcolor{black}{This subsection deals with the performance and the training of the models on CIFAR-10 dataset \cite{krizhevsky2009learning} using Genetic CFL architecture and its comparison with the performance of the generic FL architecture.} The training process of this dataset is similar to the training of the MNIST handwritten digits dataset. The server initializes the model, distributes the weights of the server model to every client device, the models are trained on the random subset of the dataset assigned for two epochs, the current hyper-parameters are subjected to genetic evolution, the trained weights are sent back to the server to get aggregated. This process is repeated for several rounds. The performance of the server model after each round, at the end of the aggregation phase is plotted in Fig.~\ref{fig:cifar_per} and tabulated in Table~\ref{cifar_cfl}.

\begin{table*}[htbp]
\caption{Performance of FL against Genetic CFL on CIFAR-10 dataset for various hyper-parameters}
\begin{center}
\begin{tabular}{|c|c|c|c|c|c|}
\hline
\multirow{2}{*}{\textbf{Client Ratio}}&\multirow{2}{*}{\textbf{Rounds}}&\multicolumn{2}{|c|}{\textbf{FL}}&\multicolumn{2}{|c|}{\textbf{Genetic CFL}} \\
\cline{3-6} 
& & \textbf{\textit{Accuracy}}& \textbf{\textit{Loss}} & \textbf{\textit{Accuracy}}& \textbf{\textit{Loss}}\\
\hline
\multirow{3}{*}{0.10} & 3 & 0.6818 & 0.9540 & 0.6514 & 1.0097\\
\cline{2-6}
& 6 & 0.6891 & 1.3746 & 0.6639 & 1.3447\\
\cline{2-6}
& 10 & 0.6862 & 1.6617 & 0.6599 & 1.6449\\
\hline
\multirow{3}{*}{0.15} & 3 & 0.6973 & 0.9612 & 0.7098 & 0.8814\\
\cline{2-6}
& 6 & 0.6988 & 1.3675 & 0.7199 & 1.693\\
\cline{2-6}
& 10 & 0.6952 & 1.6225 & 0.7129 & 1.3806\\
\hline
\multirow{3}{*}{0.30} & 3 & 0.7578 & 0.8708 & 0.7688 & 0.7818\\
\cline{2-6}
& 6 & 0.7634 & 1.0891 & 0.7646 & 0.981\\
\cline{2-6}
& 10\ & 0.7613 & 1.2964 & 0.7623 & 1.2961\\
\hline
\end{tabular}
\label{cifar_cfl}
\end{center}
\end{table*}

The performance of the models trained on the hyper-parameters that are optimized using genetic algorithm for the respective clusters is higher than those that are not. This performance is consistent with any number of client devices. The performance also improves as the client ratio increases. The lowest loss is encountered at the second round for client ratio $0.3$. The accuracy however peaks at the fourth epoch with a decent amount of loss for prediction. Any further training of the models do not provide better performance causing over-fitting. The training of the models is stopped at round two. The aggregated model therefore provides a significant performance boost for very few rounds. This provides speed and high throughput while deployment in a real time system.

\subsection{Performance Analysis of Genetic CFL}\label{sec:perform_analys}
The Genetic CFL algorithm performs better with a higher sample size. Higher number of observations per cluster should therefore improve the optimization of the hyper-parameters. However, taking into consideration the diversity of datasets both in the data characteristics and the number of data points, proper clustering of similar scenarios should provide higher throughput for the models individually. This calls for a balance between the number of clusters and the size of the cluster. A proper balance can ensure that the performance of the models in the federated architecture provides the best output in the given scenario. In a real time application the amount of edge devices expected is higher as compared to a synthetic environment. Following the progression of the performance, the higher number of total clients increases the performance significantly. The optimization of the hyper-parameters using Genetic CFL provides higher throughput for comparatively less number of rounds.\\
\begin{table*}[htbp]
\caption{Performance Comparision}
\begin{center}
\begin{tabular}{|c|c|c|c|}
\hline
\textbf{Algorithm} & \textbf{Rounds} & \textbf{MNIST} & \textbf{CIFAR-10}\\
\hline
Genetic CFL & 10 & 97.99 & 76.88\\
\hline
Byzantine Robustness of CFL\cite{sattler2020byzantine} & 200 & 97.4 & 75.3\\
\hline
FedZip\cite{malekijoo2021fedzip} & 20 & 98.03 & -\\
\hline
Iterative federated clustering\cite{ghosh2020efficient} & - & 95.25 & 81.51\\
\hline
\end{tabular}
\label{compare}
\end{center}
\end{table*}
Our architecture, Genetic CFL outperforms both algorithms \cite{sattler2020byzantine} and \cite{malekijoo2021fedzip} in accuracy and rounds. This holds up the fact that Genetic CFL architecture performs better while taking less number of rounds. In case of Iterative clustering\cite{ghosh2020efficient}, our architecture outperforms in the case of MNIST dataset but does not in the case of the CIFAR-10 data. This behaviour is attributed to the rotation and augmentation of data. This gives an upper hand in better feature extraction and representation.
Genetic optimization provides an elastic and adaptive framework for optimization of the hyper-parameters. This flexibility gives the architecture an edge over other methods by adapting to the dataset and the required environment. Most of the other types of architectures need to perform hyper-parameter tuning before hand and thus requiring manual intervention. This causes the system to be reset and a different set of parameters for a different type of data and application. this rigidity can cost both time and resources. Moreover, importance to every single client is given thus effecting not only the server model performance but also the performance of every single client device. A better delivery of service for each and every client device is ensured while increasing the performance of the server model as a whole. Table \ref{compare} shows the comparison between the performance of our architecture, Genetic CFL with other architectures that incorporate clustering in Federated Learning. The table consists of the best accuracy of the models on the MNIST handwritten digits dataset and the CIFAR-10 dataset for a given number of rounds. It is evident that the number of rounds taken is significantly less keeping the accuracy higher.

\section{Conclusion}
\label{sec:conclusion}
In this work, we have applied the genetic evolutionary algorithm to optimize the hyper-parameters - learning rate and batch size during the training of the individual end device models in a cluster for the FL architecture. We have identified and filled the gaps in the existing techniques and contributed algorithm of the Genetic CFL architecture. This architecture has been tested using MNIST handwritten digits dataset and CIFAR-10 dataset. \textcolor{black}{An accuracy of 97.99\% and 76.88\% has been respectively achieved on the datasets.} We discussed and analysed the observations and the performance of the Genetic CFL architecture. We have also covered the favourable conditions and the limitations for the algorithm to provide the best performance in deployment.\textcolor{black}{The overall performance of the models display significant rise in efficiency while reducing communication and computation cost.}\\
As part of the future work, the amount of clients and the client ratio can be scaled into larger samples closely mimicking the real time situation due to the high scalability of the model. As the population sample increases, the optimization of the hyper-parameters gets more efficient thus delivering higher throughput in the real time scenario. The type of data processed is not limited and this architecture can be used for various scenarios like Natural Language Processing tasks, Image Classification tasks and recommendation systems. Genetic CFL can also be integrated with time sensitive systems to deliver better performance in very less number of rounds.

\section*{Acknowledgement}
This manuscript is available as a preprint in Arxiv at  "https://arxiv.org/abs/2107.07233"
\bibliographystyle{hindawi_bib_style}
\bibliography{ref} 

\end{document}